# A context-aware knowledge acquisition for planning applications using ontologies

Mohannad BABLI

Universitat Politècnica de València, Department of Computer Systems and Computation, Valencia, Spain, mobab@dsic.upv.es

Eva ONAINDIA

Universitat Politècnica de València, Department of Computer Systems and Computation, Valencia, Spain, onaindia@dsic.upv.es

## Abstract

Automated planning technology has developed significantly. Designing a planning model that allows an automated agent to be capable of reacting intelligently to unexpected events in a real execution environment yet remains a challenge. This article describes a domain-independent approach to allow the agent to be context-aware of its execution environment and the task it performs, acquire new information that is guaranteed to be related and more importantly manageable, and integrate such information into its model through the use of ontologies and semantic operations to autonomously formulate new objectives, resulting in a more human-like behaviour for handling unexpected events in the context of opportunities.

## Introduction

The big improvements in Automated Planning (AP) technology open the way to address a variety of real applications such as tourism (Babli et al. 2016), travel plan generation (Knoblock et al. 2001), space technology (Guzman et al. 2015; Muscettola et al. 1998), and underwater-installation maintenance (Cashmore et al. 2018). The design of concise models for an automated agent acting in the real world has become of increasing interest in the planning community. Various tools and approaches that automate the building process of the domain models have been developed such as (Motta et al. 2004; Vaquero, Silva and Beck 2013). Designing a model of the execution environment needs to be close enough to reflect the execution environment, while on the other hand, concise enough for AP search technologies to compute solutions efficiently. The non-determinism of the environment can lead to partially developed incomplete models. Encoding all alternative situations when designing the agent domain is unreasonable and renders agents to lack autonomy. In dynamic environments where exogenous events occur and opportunities might arise without warning during plan execution, it is not possible to model the uncertainty in the world (Guzman et al. 2015; Cashmore et al. 2018) and it is computationally most efficient to plan without taking uncertainty into account. In the literature, agents monitor the execution of the plan in the environment and can formulate alternative goals on the fly based on objects that already exist in the agent model (Cox 2007; Dannenhauer and Muñoz-Avila 2013; Klenk, Molineaux and Aha 2013), or based on new objects of predefined classes (types) as the approach of opportunistic planning (Cashmore et al. 2018). Approaches in (Babli, Marzal and Onaindia 2018; Babli, Onaindia and Marzal 2018) extend models with new objects of new classes; however, the applicability of the existing action schemas to the new input object of the new class is questionable. Our motivation is to tackle this problem providing the agent from the first place with the capability of autonomously extending its planning task domain when encountering unexpected events, if and only if, such events relate to its execution environment and are manageable by its capabilities.

Our main contribution is a similarity method tailored for planning tasks which takes into consideration not only the relevance of new objects but more importantly the capabilities of the agent model to handle these objects, resulting in a domain-independent approach that extends the knowledge of a planning task. Drawing upon the richness and expressiveness of standard representations of ontologies and AP, a semantic similarity measure tailored for planning dynamics to filter out objects that are irrelevant or unmanageable by the agent, and an ontology alignment for accommodating the new acquired objects into the planning task specification, possibly triggering the formulation of a goal that induces a better-valued plan.

## Background

Consider a repair agency (*RA*) scenario; a robot in a warehouse has a task of one-day maintenance of several large kitchen appliances received by the agency. The warehouse has three areas; a transit area for items that require maintenance, an inspection area for maintenance, and a storage area for items after maintenance. The scenario is formulated as a planning task including a set of categories of large kitchen appliances, the operations that the agent can perform and their durations (movement between the warehouse areas, and a maintenance operation specialised for large kitchen appliances, loading, and unloading). The planner solves this task and returns a plan with a total repairing of three items: two of type *dishwasher* and one of type *refrigerator*. During the plan execution, the RA receives two new items *iphone_ID7500* and *bosch_ID3400* into the transit area from a different delivery agent operating in the same city, the items types are requested from that agent and are found to be *mobile_phone* and *kitchen_range*, respectively, not formerly considered in the planning task, which may represent an opportunity if the goal of repairing the new object can be aligned within the modelling of the planning task and triggers a plan compliant with the current goals, achieving an extra additional goal (the case of *bosch_ID3400*), or the agent may deem the new object as irrelevant or beyond its operational capabilities, and not erroneously integrate it in its model (the case of *iphone_ID7500*).

A *temporal planning task* is defined as $\phi = \langle \delta, \rho, \iota \rangle$, where $\delta$ is the domain which characterises the planning task behaviours (e.g., a RA domain), $\rho$ is the problem which contains the particular elements and their properties involved in the planning task $\phi$, and finally the problem instance $\iota$ in which values are assigned to the elements properties describing the initial situation and goals of $\phi$, e.g., $\rho$ and $\iota$ may represent a one-day maintenance in a RA. The elements that define the *domain* $\delta = \langle \Omega, \Psi, \Xi \rangle$ are:

- $\Omega = \{\omega_1, \omega_2, \dots\}$ is the set of classes defined in a reasonable hierarchy in $\delta$, e.g. classes of items in a RA such as *robot*, *location*, *major_appliance*, etc.
- $\Psi = \{\psi_1, \psi_2, \dots\}$ is the set of patterns in $\delta$, where each pattern represents a property associated to a class or a relationship established among classes in $\Omega$. A pattern $\psi_i \in \Psi$ is defined as: $\psi_i = \langle name(\psi_i) \ args(\psi_i) \rangle$; where the first element is the pattern name and the second element is a set of arguments that belong to $\Omega$. The second element of a pattern $\psi_i$ is a set of classes such that $args(\psi_i) \in \Omega^{|args(\psi_i)|}$, e.g., $\psi_i = \langle be \ robot \ location \rangle$, where $be$ is the pattern name and *robot* and *location* are two classes in $\Omega$.
- $\Xi = \{\xi_1, \xi_2, \dots\}$ is the set of actions schemas in $\delta$, an action schema $\xi_i \in \Xi$, e.g., $(repair \ robot \ major\_appliance \ \dots)$ is defined as $\xi_i = \langle head(\xi_i), dur(\xi_i), cond(\xi_i), eff(\xi_i) \rangle$:
  - $head(\xi_i) = \langle name(\xi_i) \ pars(\xi_i) \rangle$ is the action schema *head* defined by its name and the corresponding parameters $\{par_z\}_{z=1}^{ar(\xi_i)}$ of classes $\Omega$. As with the argument list of a pattern, $pars(\xi_i) \in \Omega^{|pars(\xi)|}$.
  - $dur(\xi_i)$ is the duration of the action schema $\xi_i$ represented by a numeric value.

- $cond(\xi_i)$ is the set of conditions of the action schema $\xi_i$ that are required at different time instances through the duration of $\xi_i$.
- $eff(\xi_i)$ is the set of positive and negative effects of the action schema $\xi_i$ that occur at different time instances through the duration of $\xi_i$.

A *problem* $\rho$ specifies the objects, $O$, involved in the problem along with a set of variables, $V$, that model the properties ($\psi$) of these objects. Formally, a problem is defined as $\rho = \langle O, V \rangle$, where:

- $O = \{o_1, o_2, \dots\}$ is the set of objects of the problem and each object is of a class in $\Omega$, e.g., items to be repaired.
- $V = \{v_1, v_2, \dots\}$ is the set of variables of the problem. A variable $v_i \in V$ is an instantiation of a pattern $\psi$ by binding the elements of $args(\psi)$ to objects in $O$. Hence, the form of a variable is $v_i = \langle name(\psi) \; \{o_i\}_{i=1}^{ar(\psi_i)} \rangle$, e.g., $v_i = \langle be \; av \; area\_inspect \rangle$ that is instantiated from the pattern $\psi_i = \langle be \; robot \; location \rangle$, where $av$ and $area\_inspect$ are objects in $O$ of classes $robot$ and $location$ respectively.

Finally, the last element of $\phi$ is the *problem instance* $\iota$, comprises two lists of pairs (`variable`,`value`), values can be `TRUE`, `FALSE`, or a `numeric`, e.g., ((`repaired refrigerator_ID03`),`TRUE`). Formally, a problem instance is defined as $\iota = \langle I, G \rangle$, where:

- $I$ represents the initial state; initially known information of variables and associated values:
  - at time instance $t = 0$, $\{(v_1,value_1),(v_2,value_2),\dots\}$; a full assignment of values to the variables of $V$ constituting the current state $s_0$.
  - plus possibly associated values at future time instances: $[t_1, \{(v_1,value_1), \dots\}]$, $[t_i, \{(v_i,value_i), \dots\}]$; the information at t=0 that is known to happen at a future time; representing exogenous happenings such as the opening and closing time of monuments, availability of tables in restaurants, the start and end operating time of a robot or driver.
- $G = \{g_1, g_2, \dots\}$ is an assignment of values to a partial set of variables in $V$ selected as goals to be accomplished by the plan computed by the planning solver.

An action $a_i$ is instantiated from an action schema $\xi_i \in \Xi$ by binding $pars(\xi_i)$ to objects in $O$, $dur(\xi_i)$ to a numeric value, $cond(\xi_i)$ to one or more pair of variables associated with positive boolean values, and $eff(\xi_i)$ to one or more pair of variables associated with positive or negative boolean values. E.g., (`move av rea_storage area_transit ...`) is an instantiation of an action (*move robot location location ...*) with ((`duration area_storage area_transit`),5) as duration, a beginning condition ((`be av area_storage`),`TRUE`), a condition through the duration of the action (⟨`active av`⟩,`TRUE`), one beginning negative effect ((`be av area_storage`),`FALSE`), and finally one end positive effect ((`be av area_transit`),`TRUE`).

The planner receives $\phi$ as input and outputs a temporal plan $\pi$ that satisfies $G$ as a collection of pairs or 2-tuple actions $\pi = \langle (t_1,a_1), (t_2,a_2), \dots \rangle$, where $t_i$ indicates the start time of the action $a_i$ which is instantiated from an action schema $\xi_i \in \Xi$ using objects defined in $O$, e.g., ⟨(`0.0003`,(`move av rea_storage area_transit`), (`5.0005`,(`load av refrigerator_ID03`), ...⟩.

## Overview of the approach

We simulate the plan execution like in a real context, using the simulation system introduced in the work of (Babli et al. 2016) which takes $\phi$ and $\pi$ as input and encodes them into a timeline as a collection of chronologically ordered timed events encapsulating the changes to be expected in the subsequent states. The monitoring process of the simulation system, on one hand, simulates 1) receiving exogenous events and adding them to the timeline, exogenous events convey external information received from other agents operating in the same environment modifying the real world states, and on the other hand, 2) the execution of each timed event, checking that conditions are successfully satisfied and the effects happen when they should, thus validating and updating, respectively, the states of the world (timeline).

Dynamically simulating plan monitoring consists in observing the state that results from executing the plan actions in the environment and checking whether the *observed state* matches the *expected state*. This operation creates a *discrepancy set* as the difference between the two sets, which will comprise variables associated with values. In this paper, we focus on discrepancies that denote a potentially achievable goal opportunity. More specifically, the discrepancy set will contain assigned variables of the form $(\langle \text{name}(\psi) \ \{o_i\}_{i=1}^{ar(\psi_i)}\rangle, \text{TRUE})$ where $\exists o \in \{o_i\} \ \wedge o \notin O$. In our approach the system requests the class $\omega$ of the new object $o$ from the source agent that delivered it, other advanced techniques can be used such as image recognition using TensorFlow in deep learning. We focus on the case when $\omega \notin \Omega$ and the agent decides whether this $\omega$ is relevant and manageable using $\Xi$ (its actions schema), integrates it, and generates a new goal that includes $o$, or on the other hand, the agent rejects the object. We designed the approach sketched in Figure 1 which works in Five stages:

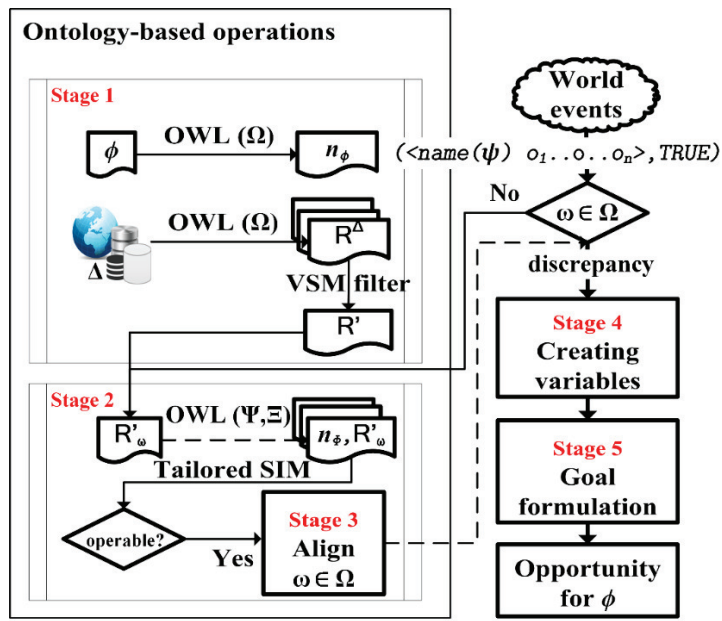

Figure 1: The ontology-based Goal formulation model

**Stage 1: Preliminary identification of similar ontologies**. First, the agent creates a preliminary OWL ontological representation called $n_\phi$ of only the classes $\Omega$ of $\phi$ ($\Omega_\phi$). Second, the system retrieves a set of partial remote planning tasks $\Delta$, specifically the classes $\Omega$ of several semi-cooperative agents from on-line repositories and creates their OWL ontological representation $R^\Delta$. Subsequently, the agent applies a preliminary quick similarity measure, vector space distance (VSM), to $R^\Delta$ and obtain the set $R'$, that contains the ontologies that may seem like $n_\phi$ according to classes.

**Stage 2: Thorough identification of similar ontologies**. When the information of a new object $o \notin O$ is received in the form of ($\langle \texttt{name}(\psi)\ o_1 \dots o \dots o_n \rangle$,`TRUE`), the agent identifies the class $\omega$ of $o$ from the source agent which delivered $o$, and finds $\omega \notin \Omega_\phi$. Second, among R′, the agent filters out ontologies that do not contain $\omega$ to get R′$_\omega$. Third, the agent extends $n_\phi$ and R′$_\omega$ to represent not only classes $\Omega$ but also patterns $\Psi$ and the heads of actions schemas $\Xi$. Hence, represented the relationship established among the agent's classes and the agent capabilities as of the operations it can perform with these classes. It is safe to assume that semi-cooperative agents would only share information related to classes $\Omega$, relationships associated with classes, and heads of operations $\Xi$ applied to classes, with no private details such as objects, goals, or states. Subsequently, the agent applies a tailored semantic similarity (TSM) measure between $n_\phi$ and ontologies in R′$_\omega$ (to consider the planning dynamics of $\Psi, \Xi$ and obtain $n_\omega$); a high similarity value means that the remote agent is equipped with similar capabilities. If the similarity value is higher than a specified threshold, then the new class $\omega$ is not only relevant but also manageable by the action's schemas $\Xi$ of $\phi$, and a low similarity value means the new class $\omega$ cannot be managed by the agent.

**Stage 3: Positioning a new class and integrating a new object**. The system attempts to position $\omega$ in $n_\phi$ via a semantic alignment with a neighbourhood constraint between $n_\phi$ and $n_\omega$. If the alignment is successful $\omega$ is positioned in the hierarchy of classes in the ontology, then $\omega$ is added to: $\Omega_\phi$, as arguments for relating $\Psi_\phi$ and as parameters in relating headers of the action schemas $\Xi_\phi$, and $o$ is added to $O$.

**Stage 4: Creating new variables**. If the new object $o$ is successfully positioned in $\phi$, the next step is to add the required planning variables $V$ that describe $o$, and the associated values besides ($\langle \texttt{name}(\psi)\ o_1 \dots o \dots o_n \rangle$,`TRUE`). The system automatically identifies the variables and associated values required for integrating $o$ in $\phi$ and adds it to $V, \iota$.

**Stage 5: Goal formulation**. If the class $\omega$ of object $o$ is a class or a sibling of a class that is involved in a goal $g \in G$, then we formulate $x$ candidate new goals that involve the newly received object $o$, where $x$ depends on the possible permutations of objects in the goal predicate; $x = 1$ if the goal has only $o$ as a parameter such as ($\langle \texttt{name}(\psi)\ o \rangle$,`TRUE`), (e.g. ($\langle \texttt{repaired bosh\_ID3400} \rangle$,`TRUE`).

## Ontology-based operations

In this section, we detail the tasks required to extend the planning task knowledge to include new objects.

*Preliminary OWL representation*

OWL has been the World Wide Web Consortium recommendation since 2004. In this section, we explain the classes OWL ontological representation for $n_\phi$ and R$^\Delta$. We utilised *OWL API* which is an open source Java API for creating, manipulating, and serialising OWL Ontologies (Horridge and Bechhofer 2011). We use snapshots from the GUI of Protégé to show visual explanations of the ontological representation.

The preliminary OWL ontological representation consists of a set $\mathcal{C}$ of concepts (OWL classes) that are used to represent $\Omega$, a set of OWL annotation properties used to describe $\mathcal{C}$. A concept $c \in \mathcal{C}$ can have one or many annotation properties. The symbol :: is used to refer to a subconcept, for instance, $c_i :: c_j$ means $c_j$ is a subconcept of $c_i$. Classes can be expressed as forming a class hierarchy. For each class in $\Omega$, an OWL concept is created in $n_\phi$ abiding the exact hierarchy.

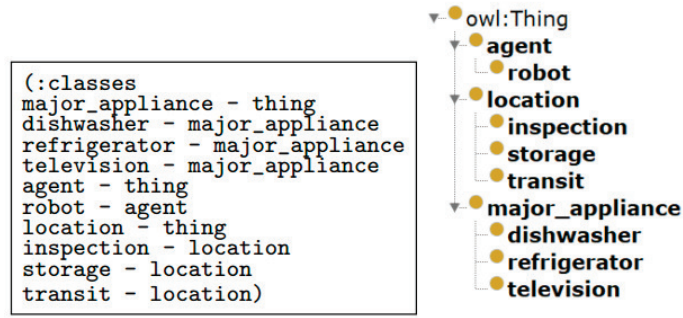

Figure 2: Representation of $\Omega_\phi$

The left part of Figure 2 shows the classes $\Omega$ of the RA $\phi$ and the right part of the figure shows the corresponding $\mathcal{C}$ in $n_\phi$. For instance, the class `dishwasher` is represented as $c_{\text{major\_appliance}}::c_{\text{dishwasher}}$.

*Preliminary identification of similar ontologies*

In ontology engineering, it is useful to know quickly if two ontologies are close or remote before deciding to match them (David and Euzenat 2008). We used a Vector Space Model similarity measure using cosine index with TF (weighted frequency term, VSM filter in Figure 1) to preliminary filter ontologies unrelated to $n_\phi$ among $R^\Delta$ (thus irrelevant classes) and obtain $R'$ as the set of ontologies that seem most similar to $n_\phi$ according to classes. VSM CosineTF has proven to obtain good results compared to other distance measures and is computed largely faster (David and Euzenat 2008). To counter the natural complication of lexical alterations, we augment the concepts of $n_\phi$ and of the ontologies $R^\Delta$ with the ConceptNet relations and concepts as OWL annotations as a standard mean to describe concepts. ConceptNet (Speer, Chin and Havasi 2017) is a knowledge graph that utilises a closed set of 36 selected relations such as isA, usedFor, hasProperty, etc., with the aim of representing concepts relationships independently of the language or the source of the terms it connects. We used an independent Java API, *OntoSim*, to compute the similarity distances between ontologies. Left part of Figure 3 shows an example of $R^\Delta = \{A, B\}$ with different terms, and right part of Figure 3 shows a small portion of the annotations attached to the concept $c_{\text{television}}$.

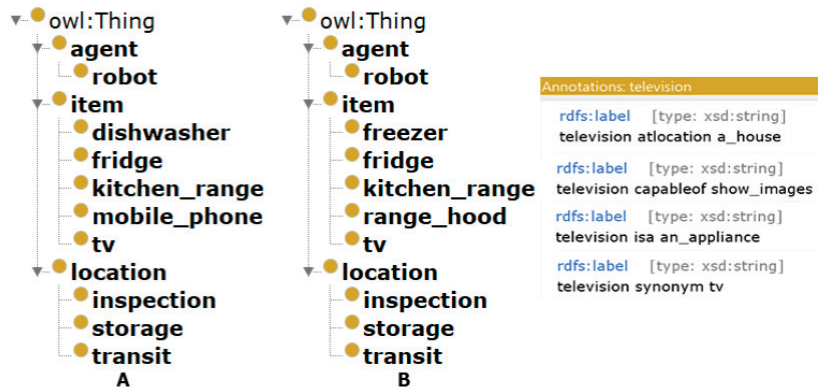

Figure 3: $R^\Delta$ with different terms and television annotations

*Extended OWL representation*

After we obtain $R'_\omega$, we extend $n_\phi$ and $R'_\omega$ utilising OWL object properties $J_{\text{hasParameter1...n}}$ to specify $args$ and $pars$ of $\Psi$ and $\Xi$ for $n_\phi$ and ontologies of $R'_\omega$. Figure 4 shows the representation of $\Psi_\phi$ and a sample of annotations for $\psi_i = \langle be$ ($dishwasher\ refrigerator\ robot\ television$) $location\rangle$, with $J_{\text{hasParameter1...2}}$ to specify $arg_1 = dishwasher \lor refrigerator \lor robot \lor television$ and $arg_2 = location$.

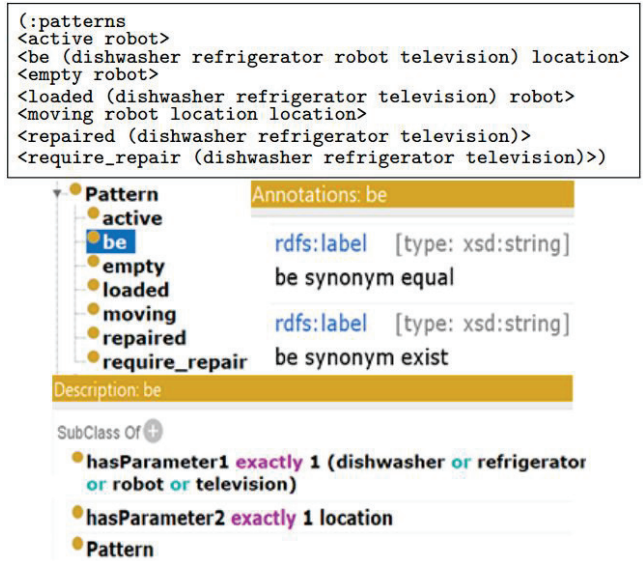

Figure 4: Representation of $\Psi_\phi$

Figure 5 shows $\Xi_\phi$ and sample of annotations for $\xi_i = \langle repair\ robot$ ($television\ refrigerator\ dishwasher$) $inspection\rangle$, with $J_{\text{hasParameter1...3}}$ to specify the parameters $robot$, ($television \lor refrigerator \lor dishwasher$), and $inspection$.

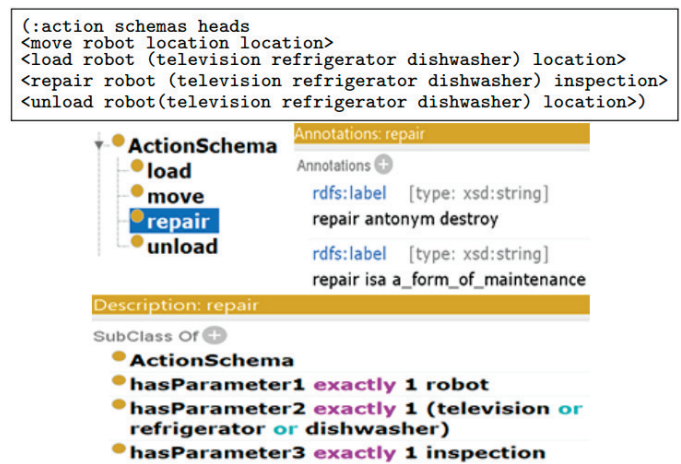

Figure 5: Representation of $\Xi_\phi$

*A tailored semantic similarity measure*

We designed Algorithm 1 (TSM algorithm) to exploit the information from the concept's names and Conceptnet annotations. Utilising a hybrid string similarity measure SoftTFIDF (Cohen, Ravikumar and Fienberg 2003) to measure similarities between concepts supporting a higher degree of syntactic variation between the terms. SoftTFIDF combines TF-IDF, a token-based similarity widely used in information retrieval (Raghavan and Wong 1986), using Jaro-Winkler edit-based distance with a threshold of 0.9 for two tokens to be considered similar to promote high precision, and with a threshold of 0.6 for the SoftTFIDF above which two concepts are considered similar.

---

**Algorithm 1** Tailored similarities for $\Omega$, $\Psi$, and $\Xi$

---

1: **if** $InflictedForm(c_1,c_2) \lor Synonyms(c_1,c_2)$ **then**
2:     `SimValue` $= 1$
3: **else if** $SoftTFIDFdistance(c_1,c_2,\text{synonym}) > 0.7$ **then**
4:     `SimValue` $= SoftTFIDFdistance(c_1,c_2,\text{synonym})$
5: **else**
6:     $Relations =$ {`synonym,isA,usedFor,atLocation,`
7:                   `capableOf,relatedTo,antonym,hasA,derivedFrom,`
8:                   `hasContext,uri}`
9:     `SimValue` $= AverageOfNonNullValues(c_1,c_2,Relations))$
10: **end if**

---

Follows an explanation of how the algorithm works on two concepts:

- If an inflected form or exact synonym, then `SimValue`=1 (lines 1,2 of Algorithm 1).
- Else if the SoftTFIDFdistance according to the synonym annotations is found to be greater than 0.7, then `SimValue`=SoftTFIDFdistance (lines 3,4 of Algorithm 1).
- Else `SimValue`= the average of the non-null values of the SoftTFIDFdistance with respect to the `Relations` (lines 5,10 of Algorithm 1).

We apply Algorithm 1 to obtain similarity matrices for classes, patterns, and heads of operators. Subsequently, the values are aggregated to obtain a final similarity value, a high value means that the newly received object $o$ of the new type $\omega$ is manageable, and an alignment can be done using the ontology with the high similarity ontology $n_\omega$, otherwise, $o$ is considered to be unmanageable even if it was considered relevant using VSM filter.

*Alignment with neighbourhood constraint*

To determine where to position the concept $c_\omega$ representing $\omega$ within the hierarchy of concepts $\mathcal{C}$ in $n_\phi$. We perform an alignment between, on one hand, $\mathcal{C}$ in $n_\phi$, and on the other hand, the particular part of the taxonomic branch of $n_\omega$ that includes $c_\omega$, the parent concept $c_{parent(\omega)}$, and the siblings $\mathcal{C}_{siblings(\omega)}$. For the alignment between the patterns of $n_\phi$ and patterns of $n_\omega$, we used the similarity matrices (produced by TSM in the previous section) with a neighbourhood constraint. Since domain-independent constraints convey general knowledge about the interaction between related nodes and perhaps the most used such constraint is the neighbourhood constraint, as suggested in (Doan et al. 2003) where "*two nodes match if nodes in their neighbourhood also match*". If $c_{parent(\omega)}$ in $n_\omega$ matches $c_x$ in $n_\phi$, then we establish $c_x :: c_\omega$ in $n_\phi$. On the other hand, if no match was achieved with the parent, the neighbourhood constraint procedure matches $\mathcal{C}$ in $n_\phi$ with $\mathcal{C}_{siblings(\omega)}$ in $n_\omega$, if the matching siblings percentage exceeds a specified threshold, and matched concepts are found to be under a common parent in $n_\phi$, then we list $c_\omega$ as a subconcept of that superconcept in $n_\phi$. If the alignment was successful, we add $\omega$ to $\Omega_\phi$ and $o$ to $O_\phi$.

## Opportunity identification

Once the $x$ candidate new goals are formulated as explained in Stage 5 of the overview of our approach, the system generates $\Phi' = \phi'_1, \ldots, \phi'_x$ (modified versions of $\phi$), where the added information includes $o$, $\omega$, the information of $o$, the discrepancy variables, $G' = g'_i \cup G$, and the new current state. We use a planner to solve each $\phi'_i \in \Phi'$ to know which $g'_i$ can be considered an opportunity to $\phi$ in the context of $\pi$. $g'_i$ is considered an opportunity goal for $\phi$, when the planner can find a plan $\pi'_i$ to solve $\phi'_i$ that includes the new goal plus the original set of goals.

## Cases of study

The aim of this section is to show the behaviour of our system and how it outperforms the approach in (Babli, Marzal and Onaindia 2018) on its application domain ($RA$). Consider the ($RA$) scenario described previously as $\phi$. The robot `av` can load one appliance at a time. Initially, the system has a set of large kitchen appliances categories in $\Omega$ as shown in the left part of Figure 2. On the other hand, $\Psi$ and $\Xi$ are shown in the upper part of Figures 4 and 5, respectively. The information of $\rho$ also includes the three areas of the warehouse (`area_transit`), (`area_inspect`), and (`area_storage`). The information of $I$ includes `av` start location ((`be av area_storage`),TRUE), state ((`empty av`),TRUE), operational hours between 10:00 and 23:00, the appliance that require repairing and their locations, and the durations of movement between the warehouse areas, maintenance time, loading time, and unloading time). The goals are to repair and deliver `washer_ID02`, `washer_ID101`, and `refrigerator_ID03` to `area_storage`. The simulator that we are using does not deal with conditional effects or derived predicates, therefore we added a dummy action with a zero duration that asserts that an item is delivered if it is repaired and it is at the storage area.

```
0.0003:    (MOVE AV AREA_STORAGE AREA_TRANSIT) [5.0000]
5.0005:    (LOAD AV WASHER_ID02 AREA_TRANSIT) [1.0000]
6.0008:    (MOVE AV AREA_TRANSIT AREA_INSPECT) [5.0000]
11.0010:   (UNLOAD AV WASHER_ID02 AREA_INSPECT) [1.0000]
12.0013:   (REPAIR AV WASHER_ID02 AREA_INSPECT) [80.0000]
91.0015:   (LOAD AV WASHER_ID02 AREA_INSPECT) [1.0000]
92.0017:   (MOVE AV AREA_INSPECT AREA_STORAGE) [5.0000]
97.0020:   (UNLOAD AV WASHER_ID02 AREA_STORAGE) [1.0000]
 98.0023:  (DUMMY WASHER_ID02 AREA_STORAGE) [0.0000]
98.0025:   (MOVE AV AREA_STORAGE AREA_TRANSIT) [5.0000]
103.0027:  (LOAD AV WASHER_ID101 AREA_TRANSIT) [1.0000]
104.0030:  (MOVE AV AREA_TRANSIT AREA_INSPECT) [1.0000]
109.0033:  (UNLOAD AV WASHER_ID101 AREA_INSPECT) [1.0000]
110.0035:  (REPAIR AV WASHER_ID101 AREA_INSPECT) [80.0000]
189.0038:  (LOAD AV WASHER_ID101 AREA_INSPECT) [1.0000]
190.0040:  (MOVE AV AREA_INSPECT AREA_STORAGE) [5.0000]
195.0043:  (UNLOAD AV WASHER_ID101 AREA_STORAGE) [1.0000]
 196.0045: (DUMMY WASHER_ID101 AREA_STORAGE) [0.0000]
196.0047:  (MOVE AV AREA_STORAGE AREA_TRANSIT) [5.0000]
201.0050:  (LOAD AV REFRIGERATOR_ID03 AREA_TRANSIT) [1.0000]
202.0052:  (MOVE AV AREA_TRANSIT AREA_INSPECT) [5.0000]
207.0055:  (UNLOAD AV REFRIGERATOR_ID03 AREA_INSPECT) [1.0000]
208.0058:  (REPAIR AV REFRIGERATOR_ID03 AREA_INSPECT) [80.0000]
287.0060:  (LOAD AV REFRIGERATOR_ID03 AREA_INSPECT) [1.0000]
288.0063:  (MOVE AV AREA_INSPECT AREA_STORAGE) [5.0000]
293.0065:  (UNLOAD AV REFRIGERATOR_ID03 AREA_STORAGE) [1.0000]
196.0045:  (DUMMY REFRIGERATOR_ID03 AREA_STORAGE) [0.0000]
```

Figure 6: PLAN1

The plan to solve $\phi$ (PLAN1 in Figure 6) is calculated by the planner and consists of 27 actions; the robot `av` moves from its start location (`area_storage`) to (`area_transit`), loads an item, moves to (`area_inspect`), unloads the item, repairs the item, loads the item, moves to (`area_storage`), unloads the item, and the item is delivered; the previous rotation (8+1 dummy) applies for each of the three appliances that require repairing.

The simulator starts PLAN1 execution simulation. Let us assume that after repairing and delivering the appliance `washer_ID02` (after the ninth action in PLAN1, Figure 6), new information is received from a different delivery agent ((`be iphone_ID7500 area_transit`),TRUE) that includes the new object `iphone_ID7500` $\notin \mathcal{O}$. The system requests and finds $\omega$=`mobile_phone` $\notin \Omega$. The system creates preliminary $n_\phi$ (right part of Figure 2). Second, the system retrieves $\Omega$ of remote planning tasks $\Delta = \{A, B, C\}$ (Figures 7, 8, and 9, respectively) from on-line repositories.

```
● (:classes
  agent item location - thing
  robot - agent
  dishwasher fridge kitchen_range mobile_phone tv - item
  inspection storage transit - location)

● (:patterns
<be (dishwasher fridge kitchen_range mobile_phone robot tv) location>
<loaded (dishwasher fridge kitchen_range mobile_phone tv) robot>
<repaired (dishwasher fridge kitchen_range mobile_phone tv)>
<require_repair (dishwasher fridge kitchen_range mobile_phone tv))>
<motherboard mobile_phone>
<keyboard mobile_phone>
<battery mobile_phone>

● (:action schemas heads
<load robot (dishwasher fridge kitchen_range mobile_phone tv) location>
<repair2 robot mobile_phone inspection>
<unload robot(dishwasher fridge kitchen_range mobile_phone tv) location>)
```

Figure 7: $\Omega$, $\Psi$, and $\Xi$ of remote planning task $A$

```
● (:classes
  agent item location - thing
  robot - agent
  freezer fridge kitchen_range range_hood tv - item
  inspection storage transit - location)

● (:patterns
<be (dishwasher fridge freezer fridge kitchen_range range_hood tv robot) location>
<loaded (freezer fridge kitchen_range range_hood tv) robot>
<repaired (freezer fridge kitchen_range range_hood tv)>
<require_repair (freezer fridge kitchen_range tv))>

● (:action schemas heads
<load robot (freezer fridge kitchen_range range_hood tv) location>
<fix robot (freezer fridge kitchen_range range_hood tv) inspection>
<unload robot (freezer fridge_kitchen_range range_hood tv) location>)
```

Figure 8: $\Omega$, $\Psi$, and $\Xi$ of remote planning task $B$

```
● (:classes
  person accommodation attraction restaurant - thing
  hotel - accommodation
  aquarium zoo tower architectural_building park cathedral - attraction)
```

Figure 9: $\Omega$ of remote planning task $C$

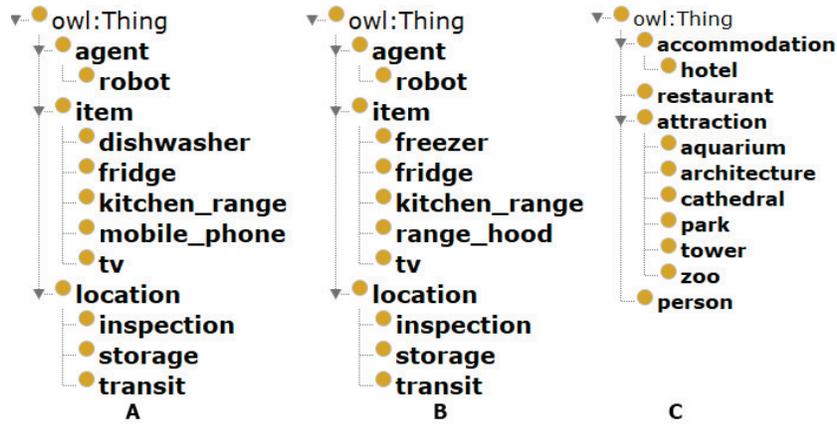

Figure 10: R$^\Delta$ preliminary representation

Then creates their preliminary representation R$^\Delta$ (Shown in of Figure 10). The ontologies are augmented using ConceptNet. VSM distance is calculated between $n_\phi$, and each ontology in R$^\Delta$, the distances are respectively: 0.79, 0.78, 0.38. The system tries to find *mobile_phone* in R$^\Delta$ and creates R′$_\omega$ = {A}, the only remote ontology that contains the class $\omega$ and with high similarity value (0.79) to $n_\phi$. If $A$ was to be recognised as $n_\omega$ and used for the alignment, like in (Babli, Marzal and Onaindia 2018), the result would be erroneously integrating `iphone_ID7500` and generating a goal the agent cannot manage. Instead, in our approach, we extend the representation of $n_\phi$ and $A$, (samples for extending $n_\phi$ are shown in Figures 4 and 5) and apply the TSM distance, the distance is found 0.39 and the agent deems `iphone_ID7500` unmanageable.

The simulation of PLAN1 continues. After `av` has repaired washer_ID101, new information is received (⟨be `bosch_ID3400 area_transit`⟩,TRUE) that includes the new object `bosch_ID3400` ∉ $O$ of class $\omega$=`kitchen_range` ∉ Ω. The system creates R′$_\omega$ = {A, B}, and the TSM distances with respect to $n_\phi$ are respectively: 0.39, 0.64. Subsequently, $B$ is used to position $c_{kitchen\_range}$ in $n_\phi$ by applying alignment with neighbourhood constraint, the system finds that the matching percentage is 66% according to siblings ($C_{siblings(c_{kitchen\_range})}$ shown in Figure 3), and therefore positions `kitchen_range` in $n_\phi$ as $c_{major\_appliance}$ :: $c_{kitchen\_range}$. A new entry `kitchen_range - major_appliance` is added to Ω, and as *args* and *pars* for relating Ψ and Ξ, `bosch_ID3400` is added to $O$. The information required for integrating `bosch_ID3400` in $\phi$ is automatically identified, requested, and added to $V$ and $\iota$. Since kitchen_range is a sibling of a type that is involved in a goal $g \in \mathcal{G}$ thus, the system formulates new $g′_1$ = (⟨repaired `bosch_ID3400`⟩,TRUE), $g′_2$ = (⟨delivered `bosch_ID3400`⟩,TRUE) and $\mathcal{G}′$ = $g′_1 \cup g′_2 \cup \mathcal{G}$. The system updates the current state at the time the new information was received. The planner is called to generate a new plan (PLAN2 shown in Figure 11); allowing the robot to repair and deliver the original set of items plus the new item. At the end of the day, the robot ends up in repairing 4 items instead of 3.

```
196.0047:    (MOVE AV AREA_STORAGE AREA_TRANSIT)
201.0050:    (LOAD AV REFRIGERATOR_ID03 AREA_TRANSIT)
202.0052:    (MOVE AV AREA_TRANSIT AREA_INSPECT)
207.0055:    (UNLOAD AV REFRIGERATOR_ID03 AREA_INSPECT)
208.0058:    (REPAIR AV REFRIGERATOR_ID03 AREA_INSPECT)
287.0060:    (LOAD AV REFRIGERATOR_ID03 AREA_INSPECT)
288.0063:    (MOVE AV AREA_INSPECT AREA_STORAGE)
293.0065:    (UNLOAD AV REFRIGERATOR_ID03 AREA_STORAGE)
196.0045:    (DUMMY REFRIGERATOR_ID03 AREA_STORAGE)
196.0047:    (MOVE AV AREA_STORAGE AREA_TRANSIT)
201.0050:    (LOAD AV BOSCH_ID3400 AREA_TRANSIT)
202.0052:    (MOVE AV AREA_TRANSIT AREA_INSPECT)
207.0055:    (UNLOAD AV BOSCH_ID3400 AREA_INSPECT)
208.0058:    (REPAIR AV BOSCH_ID3400 AREA_INSPECT)
287.0060:    (LOAD AV BOSCH_ID3400 AREA_INSPECT)
288.0063:    (MOVE AV AREA_INSPECT AREA_STORAGE)
293.0065:    (UNLOAD AV BOSCH_ID3400 AREA_STORAGE)
294.0067:    (DUMMY BOSCH_ID3400 AREA_STORAGE)
```

Figure 11: PLAN2

## Conclusion

Context awareness is crucial for any intelligent agent that operates in a dynamic environment. In this paper we have presented a similarity method tailored for planning tasks with a domain-independent approach that may be considered as a context model in an ambient intelligent planning service to accommodate only relevant and manageable new information on the fly during the execution of the initial plan that solves the original planning task, that in turn may trigger the formulation of new goals and produce new plans to achieve the new goals in addition to the original set of goals. The advantages of our approach are that it is domain-independent and requires only the planning task of the agent and the original plan as input, no prior-knowledge of exogenous events of possible opportunities is required as the agent has the autonomous capabilities to extend its knowledge. To do so, the system makes use of the public planning information shared by semi-cooperative agents and translate this information into ontologies. On the other hand, if agents would share an ontological representation of their public planning information, that would boost the performance; however, that is a gap between the planning community and the knowledge community which is being bridged by recent research efforts. For future work, we intend to endow the system with the ability to perform failure prediction and plan repair intelligently without delegating that task to a planner.

## Acknowledgement

This work is supported by the Spanish MINECO project TIN2017-88476-C2-1-R.